\documentclass[letterpaper]{article} 
\usepackage{aaai25}  
\usepackage{times}  
\usepackage{helvet}  
\usepackage{courier}  
\usepackage[hyphens]{url}  
\usepackage{graphicx} 
\usepackage{natbib}  
\usepackage{caption} 
\usepackage{algorithm}
\usepackage{algorithmic}

\usepackage{newfloat}
\usepackage{listings}

\usepackage{amsmath}
\usepackage{amsfonts}     
\usepackage{amssymb}      
\usepackage{booktabs}
\usepackage{multirow}
\usepackage{xcolor}
\usepackage{pifont}

\DeclareCaptionStyle{ruled}{labelfont=normalfont,labelsep=colon,strut=off} 
\floatstyle{ruled}
\newfloat{listing}{tb}{lst}{}
\floatname{listing}{Listing}
\title{Can We Get Rid of Handcrafted Feature Extractors? \\ SparseViT: Nonsemantics-Centered, Parameter-Efficient Image Manipulation Localization through Spare-Coding Transformer}
\author{
    Lei Su\textsuperscript{\rm 1, \rm 2},
    Xiaochen Ma\textsuperscript{\rm 3},
    Xuekang Zhu\textsuperscript{\rm 1, \rm 2},
    Chaoqun Niu\textsuperscript{\rm 1, \rm 2},
    Zeyu Lei\textsuperscript{\rm 1, \rm 2, \rm 4},
    Ji-Zhe Zhou\textsuperscript{\rm 1, \rm 2}\thanks{Corresponding Author: Ji-Zhe Zhou, jzzhou@scu.edu.cn}\\
}
\affiliations{
    \textsuperscript{\rm 1}College of Computer Science, Sichuan University, China\\
     \textsuperscript{\rm 2}Engineering Research Center of Machine Learning and industry inteligence, Ministry of Education of China\\
     \textsuperscript{\rm 3}Mohamed Bin Zayed University of Artificial Intelligence\\
     \textsuperscript{\rm 4}Department of Computer and Information Science, University of Macao, Macao SAR\\

}

\usepackage{bibentry}

\begin{document}

\maketitle

\begin{abstract}
Non-semantic features or semantic-agnostic features, which are irrelevant to image context but sensitive to image manipulations, are recognized as evidential to Image Manipulation Localization (IML). Since manual labels are impossible, existing works rely on handcrafted methods to extract non-semantic features. Handcrafted non-semantic features jeopardize IML model's generalization ability in unseen or complex scenarios. Therefore, for IML, the elephant in the room is:~\textbf{How to adaptively extract non-semantic features?} Non-semantic features are context-irrelevant and manipulation-sensitive. That is, within an image, they are consistent across patches unless manipulation occurs. Then, spare and discrete interactions among image patches are sufficient for extracting non-semantic features. However, image semantics vary drastically on different patches, requiring dense and continuous interactions among image patches for learning semantic representations. Hence, in this paper, we propose a Sparse Vision Transformer (SparseViT), which reformulates the dense, global self-attention in ViT into a sparse, discrete manner. Such sparse self-attention breaks image semantics and forces SparseViT to adaptively extract non-semantic features for images. Besides, compared with existing IML models, the sparse self-attention mechanism largely reduced the model size (max 80\% in FLOPs), achieving stunning parameter efficiency and computation reduction. Extensive experiments demonstrate that, without any handcrafted feature extractors, SparseViT is superior in both generalization and efficiency across benchmark datasets.
\end{abstract}


\begin{links}
    \link{Code}{https://github.com/scu-zjz/SparseViT}
\end{links}

\section{Introduction}

With the rapid development of image editing tools and image generation technologies, image manipulation has become exceedingly convenient. To address this trend, researchers have developed Image Manipulation Localization (IML) techniques to identify specific manipulated regions within images. Due to the inevitable artifacts (manipulation traces) left on an image after manipulation, these artifacts can be divided into semantic and non-semantic (Semantic-Agnostic) features. Semantic-Agnostic Features refer to features that highlight low-level artifacts information, which are independent of the image’s semantic content. These features show significant differences in distribution between manipulated and unmanipulated regions of an image. \cite{guillaro2023trufor} Existing backbone networks \cite{simonyan2014very} \cite{wang2020deep} \cite{dosovitskiy2020image16}, primarily designed for semantic-related tasks, are effective at extracting the semantic features of manipulated images. For extracting non-semantic features, most existing methods rely on handcrafted feature extractors \cite{zhou2018learningrich} \cite{bayar2018constrained} \cite{cozzolino2019noiseprint}. As shown in Table \ref{tab:model_information}, almost all existing IML models follow a design of "semantic segmentation backbone network" combined with "handcrafted non-semantic feature extraction."

\begin{table}[t]
\centering
\setlength{\tabcolsep}{2pt}
\begin{tabular}{@{}cccc@{}}
\toprule
\textbf{Model Name}  & \textbf{Backbone}  & \textbf{Extractor}           & \textbf{Manual} \\ \midrule
ManTraNet $\textcolor{gray}{\textit{\textsubscript{(CVPR19)}}}$                & VGG                & Bayar+SRM                 & \ding{51}
                       \\
SPAN $\textcolor{gray}{\textit{\textsubscript{(ECCV20)}}}$                 & VGG              & Bayar+SRM & \ding{51}                      \\
MVSS $\textcolor{gray}{\textit{\textsubscript{(ICCV21)}}}$                   & ResNet-50          &  Bayar+Sobel             & \ding{51}                       \\
CAT-Net $\textcolor{gray}{\textit{\textsubscript{(IJCV22)}}}$                     & HRNet              & DCT                     & \ding{51}                        \\
ObjectFormer $\textcolor{gray}{\textit{\textsubscript{(CVPR22)}}}$            & CNN+ViT    & DCT                    & \ding{51}                        \\
NCL-IML $\textcolor{gray}{\textit{\textsubscript{(ICCV23)}}}$                     & ResNet-101         & Sobel                 & \ding{51}                        \\
TruFor $\textcolor{gray}{\textit{\textsubscript{(CVPR23)}}}$                    & ViT & Noiseprint                 & \ding{51}                       \\ \bottomrule
\end{tabular}
\caption{Comparison and Summary of IML Models. We have indicated whether these methods rely on handcrafted feature extraction and specified the type of extractors used.}
\label{tab:model_information}
\end{table}

However, this approach requires custom extraction strategies for different non-semantic features, lacking adaptability in extracting these features. Consequently, this method is limited in improving the model's ability to adapt to unknown scenarios. Unlike traditional methods that manually extract non-semantic features, we propose an adaptive mechanism to extract non-semantic features in manipulated images. We recognize that the semantic features of an image exhibit strong continuity and significant contextual correlation \cite{wang2018non}, meaning that local semantic features are often inadequate in representing the global semantics of the image. Thus, tight and continuous interactions between local regions are necessary to construct global semantic features. In contrast, the non-semantic features of an image, such as frequency and noise, are highly sensitive to manipulation and show greater independence across different regions of the image. This characteristic allows us to employ sparse coding to establish global interactions for non-semantic features, utilizing their sensitivity to detect manipulations.

Based on this concept, we introduce SparseViT, a novel Sparse Vision Transformer. SparseViT employs a sparse self-attention mechanism, redesigning the dense, global self-attention in ViT to better adapt to the statistical properties of non-semantic features. Through sparse processing, the self-attention mechanism selectively suppresses the expression of semantic information, focusing on capturing non-semantic features related to image manipulation. Using a hierarchical strategy, SparseViT applies varying degrees of sparsity at different levels to finely extract non-semantic features. We also designed a multi-scale fusion module (LFF) as the decoder, which integrates feature maps extracted at different sparsity levels, enriching the model’s understanding of non-semantic content across multiple scales and enhancing its robustness. This design enables SparseViT to focus on learning manipulation-sensitive non-semantic features while ignoring semantic features, allowing for adaptive extraction of non-semantic features from images.

To our knowledge, there are currently no models explicitly designed for adaptive extraction of non-semantic features. SparseViT can be considered a pioneering work in adaptive extraction of non-semantic features. All our experiments were conducted under the same evaluation protocol. All models were trained on the CAT-Net \cite{kwon2021cat} dataset and tested on multiple benchmark datasets. Our proposed method demonstrated outstanding image manipulation localization capabilities across several benchmark datasets, with our model achieving the best average performance compared to others. In summary, our contributions are as follows: 
\begin{itemize}
    \item We reveal that semantic features in an image require continuous local interactions to construct global semantics, while non-semantic features, due to their local independence, can achieve global interactions through sparse encoding.
    \item Based on the distinct behaviors of semantic and non-semantic features, we propose using a sparse self-attention mechanism to adaptively extract non-semantic features from images.
    \item To address the non-learnability of traditional multi-scale fusion methods, we introduce a learnable multi-scale supervision mechanism. 
    \item Our proposed SparseViT maintains parameter efficiency without relying on feature extractors and achieves state-of-the-art (SoTA) performance and excellent model generalization capabilities across four public datasets.
\end{itemize}

\section{Related Work}

\subsection{Artifacts Extraction}

Early image manipulation localization methods primarily relied on handcrafted convolutional kernels to extract non-semantic features from images. For example, BayarConv \cite{bayar2018constrained} designed a convolutional kernel with a high-pass filter structure to capture noise patterns in images. RGB-N \cite{zhou2018learningrich} introduced SRM filters to capture differences in noise distribution, thereby representing non-semantic features. With the success of deep learning in various computer vision and image processing tasks, many recent techniques have also adopted deep learning to address image manipulation localization \cite{zhou2018learningrich}. However, due to the limitations of existing networks designed for semantic-related tasks in representing non-semantic features, 
nearly all manipulation localization methods currently rely on semantic segmentation backbone networks combined with handcrafted non-semantic feature extraction.

For instance, ManTra-Net \cite{wu2019mantra} and SPAN \cite{hu2020span} both integrate BayarConv and SRM as the first layer of their models. ObjectFormer \cite{wang2022objectformer}, based on the Transformer architecture, additionally employs a handcrafted DCT module to extract high-frequency features, enabling better capture of non-semantic characteristics in images. TruFor \cite{guillaro2023trufor} uses the handcrafted Noiseprint \cite{cozzolino2019noiseprint} feature extractor and, through contrastive learning, leverages these extracted features to enhance its manipulation detection and localization capabilities. NCL \cite{ncl-iml}  utilizes a Sobel-based \cite{dong2022mvss} non-semantic feature extractor to enhance its capability in identifying non-semantic features.
The methods for extracting non-semantic features from manipulated images by each model are shown in Table \ref{tab:model_information}.

\subsection{Sparse Self-Attention in Vision Transformers}

The Transformer was initially proposed to address natural language processing (NLP) tasks and was first applied to sequence data. The paper \cite{dosovitskiy2020image16} introduced a novel Vision Transformer (ViT) model, providing new insights for applying Transformers to the visual domain.

\begin{figure*}[t]
\centering
\includegraphics[width=\textwidth]{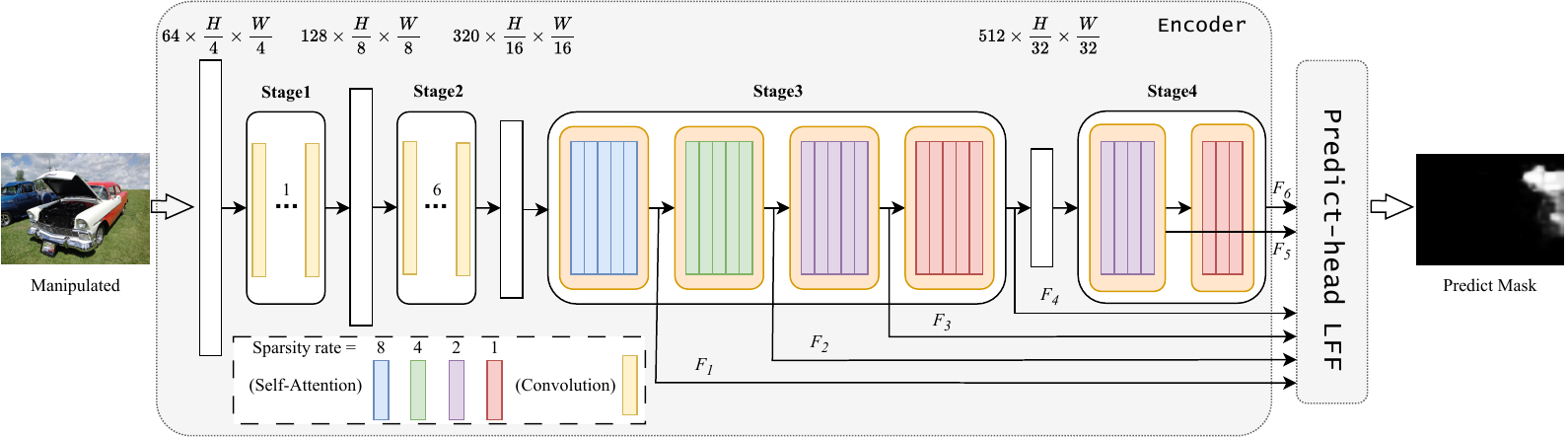} 
\caption{SparseViT. SparseViT consists of two key components: an encoder with a sparse self-attention mechanism and a prediction head (LFF) for multi-scale feature fusion. More detailed information about each module will be presented in Method.}
\label{SparseViT}
\end{figure*}
Since the introduction of Transformers in the visual domain, research on sparse attention has never ceased. The Swin Transformer \cite{liu2021swin} aggregates attention using shifted windows within a hierarchical structure. The Sparse Transformer \cite{child2019generating} reduces computational complexity by limiting the number of non-zero elements in the attention weights. ResMLP \cite{touvron2022resmlp} incorporate local connections into the attention mechanism, while \cite{liu2021pay} utilizes the non-linear properties of MLPs to replace traditional attention computation. ViViT \cite{arnab2021vivit} and CSWin Transformer \cite{dong2022cswin} reduce computational cost and improve the model's ability to handle long sequences by decomposing multi-head self-attention within the transformer. ViViT decomposes attention into temporal and spatial calculations, while CSWin Transformer splits multi-head self-attention into two parallel groups, one handling horizontal stripes and the other handling vertical stripes, forming a cross-shaped window. Focal Self-attention \cite{yang2021focal} sparsifies the attention pattern by combining fine-grained local and coarse-grained global interactions. In the field of IML, no current method has proposed using sparse attention to adaptively extract semantic-agnostic information from manipulated images. Our work is pioneering in the IML field.

\section{Method}
Manipulated instances in current datasets often focus on operations such as moving, deleting, or copying entire objects. This allows existing models \cite{pun2015image} to identify manipulated regions relatively well by relying solely on semantic features. However, this over-reliance on semantic features neglects the importance of non-semantic features, limiting the model's generalization ability in unfamiliar or complex manipulation scenarios. We observe that an image's semantic information exhibits strong continuity and contextual dependency \cite{wang2018non}, necessitating global attention mechanisms to reinforce interactions between local and global regions \cite{vaswani2017attention}. In contrast, non-semantic information tends to remain consistent between local and global features and demonstrates greater independence across different regions of an image \cite{ulyanov2018deep}. By leveraging this distinction, we can design a mechanism that reduces reliance on semantic information while enhancing the capture of non-semantic information.

To this end, we propose decomposing the global attention mechanism into a ``sparse attention'' form. Sparse attention, when representing an image's semantic information, prevents the model from overfitting to it, allowing the model to focus more on non-semantic information in the image. As shown in Figure \ref{SparseViT}, we have improved the traditional attention calculation in Uniformer \cite{li2023uniformer} by replacing global self-attention with sparsely self-attention, featuring an exponential decay in sparsity.

\subsection{Sparse Self-Attention}
Traditional deep models focus on detecting semantic objects, aiming to fit these semantic objects. Consequently, traditional self-attention employs a global interaction mode, where every patch in the image participates in token-to-token attention computation with all other patches \cite{liu2021swin} \cite{yuan2021tokens}. However, in the domain of image manipulation localization, such global interactions introduce numerous irrelevant key-value pairs. Moreover, the model's overemphasis on semantic information means that during global interaction, it takes into account features of all patches in the image, such as color and shape, leading to a comprehensive understanding of the image's overall content. Since the model primarily focuses on the overall semantic structure of the image during global interaction, it tends to overlook the local inconsistencies in non-semantic information that arise after manipulation.

\begin{figure}[t]
\centering
\includegraphics[width=0.98\columnwidth]{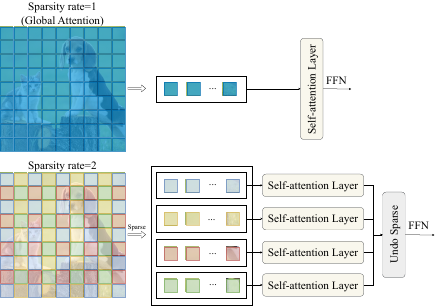} 
\caption{Sparse Self-Attention. A diagram illustrating the calculation of sparse attention. The self-attention computation occurs only between image patches of the same color.}
\label{Sparse_and_globle_attention}
\end{figure}

To address this issue, we propose using sparse attention to replace the original global attention. We introduce a new architectural hyperparameter called the ``sparsity rate'', abbreviated as ``$\mathcal{S}$''. Given an input feature map $X \in \mathbb{R}^{H \times W \times C}$, instead of applying attention to the entire $H \times W$ feature map, we divide the features into tensor blocks with a shape of $(\mathcal{S} \times \mathcal{S}, \frac{H}{\mathcal{S}} \times \frac{W}{\mathcal{S}}, C)$. This means that the feature map is decomposed into $\mathcal{S} \times \mathcal{S}$ non-overlapping tensor blocks of size $\frac{H}{\mathcal{S}} \times \frac{W}{\mathcal{S}}$, and self-attention computation is performed within these tensor blocks separately. As illustrated in the Figure \ref{Sparse_and_globle_attention}, only tensor blocks marked with the same color will perform self-attention computations. This design suppresses the expression of semantic information in sparse attention blocks, allowing the model to focus on extracting non-semantic features. Additionally, the sparsification of tensor blocks in the feature map eliminates the need for attention calculations involving a large number of irrelevant key-value pairs in manipulation localization, thereby reducing FLOPs.


\subsection{Multi-scale Features}
In the task of image manipulation localization, introducing multi-scale supervision with varying sparsity rates is crucial. Feature maps with smaller sparsity rates are rich in semantic information, which helps the model understand the global context and structure of the image. Conversely, feature maps with larger sparsity rates contain more non-semantic information, aiding the model in capturing image details and local features. This introduction of multi-scale supervision allows the model to adaptively extract various non-semantic features by suppressing semantic features to different extents, thereby enhancing its generalization ability across different visual scenes. 

As shown in Figure \ref{SparseViT}, we introduce different sparsity rates in various blocks of Stage 3 and Stage 4. The calculation method for the sparsity rates of each block in Stage 3 and Stage 4 is as follows:

\begin{equation}
S3_{\mathcal{S}}^{b_i} = 2^{\left(3 - \frac{i}{5}\right)}, \quad i = 0 \ldots 19
\end{equation}
\begin{equation}
S4_{\mathcal{S}}^{b_i} = 2^{\left(1 - \frac{i}{4}\right)}, \quad i = 0 \ldots 6
\end{equation}
Here, the superscript $b_i$ represents different layers within a Stage, where each layer is numbered starting from 0, and the subscript $\mathcal{S}$ indicates sparsity. We use the output of the last block in Stage 3 and Stage 4 at different sparsity rates as our multi-scale feature maps. Additionally, due to the sparsification of global attention, we can easily obtain multi-scale information. This approach not only significantly improves the model's accuracy and performance without increasing computational burden but also makes the model more efficient and robust.

\begin{table*}[t]
\centering
\setlength{\tabcolsep}{4pt}
\begin{tabular}{@{}ccccccccccccc@{}}
\toprule
\multirow{2}{*}{\textbf{Version}} & \multirow{2}{*}{\textbf{Parameter}} & \multirow{2}{*}{\textbf{FLOPs}} & \multicolumn{2}{c}{\textbf{COVERAGE}} & \multicolumn{2}{c}{\textbf{Columbia}} & \multicolumn{2}{c}{\textbf{CASIAv1}} & \multicolumn{2}{c}{\textbf{NIST16}} & \multicolumn{2}{c}{\textbf{DEF-12k}} \\ \cmidrule(l){4-13} 
                                 
&                                     &                                 & \textbf{F1}        & \textbf{AUC}      & \textbf{F1}       & \textbf{AUC}      & \textbf{F1}      & \textbf{AUC}      & \textbf{F1}      & \textbf{AUC}     & \textbf{F1}      & \textbf{AUC}      \\ \midrule
Uniformer                         &1.00                                   &1.00                                & 0.378                 & 0.911                 & 0.873                 & 0.938                 & 0.789                & 0.971                 & 0.326                & 0.848                & 0.182                & 0.810                 \\
Uniformer (Sparse)               & 1.00                                   & 0.83                               & \underline{0.485}                 & \underline{0.925}                 & 0.936                 & \textbf{0.976}                 & \underline{0.813}                & \underline{0.978}                 & \underline{0.355}                & \underline{0.857}                & 0.187                & 0.809                 \\
Uniformer (LFF)                    & 1.01                                    & 1.01                               & 0.473                 & 0.923                 & 0.916                 & 0.945                 & 0.808                & 0.972                 & 0.338                & 0.850                & \textbf{0.202}                & \underline{0.811}                 \\
Uniformer (Sobel+LFF)              & 1.05                                   & 1.01                               & 0.312                 & 0.882                 & 0.918                 & 0.950                 & 0.768                & 0.962                 & 0.313                & 0.845                & 0.136                & 0.755                 \\
Uniformer (Bayar+LFF)              & 1.05                                   & 1.01                               & 0.375                 & 0.909                 & 0.891                 & 0.919                 & 0.795                & 0.971                 & 0.322                & 0.853                & 0.148                & 0.778                 \\
Uniformer (DCT+LFF)                & 1.05                                   & 1.03                               & 0.472                 & 0.922                 & \underline{0.937}                 & \underline{0.974}                 & 0.805                & 0.974                 & 0.336                & 0.851                & 0.173                & 0.800                 \\
Uniformer (SRM+LFF)                & 1.05                                   & 1.02                               & 0.457                 & 0.919                 & 0.930                 & 0.953                 & 0.793                & 0.969                 & 0.332                & 0.841                & 0.189                & 0.780                 \\
SparseViT (Sparse+LFF)            & 1.01                                   & 0.84                               & \textbf{0.513}                 & \textbf{0.935}                 & \textbf{0.959}                 & 0.970                 & \textbf{0.827}                & \textbf{0.982}                & \textbf{0.384}               & \textbf{0.861}                & \underline{0.197}                & \textbf{0.816}                 \\ \bottomrule
\end{tabular}
\caption{Ablation Study of SparseViT. Trained on the CAT-Net joint dataset and validated on CASIAv1. The number of parameters and FLOPs are expressed as multiples of the backbone Uniformer. The best numbers in each column are highlighted in bold, and the second-ranked
results are underlined. The consistently improving performance demonstrates the necessity of Sparse and LFF.}
\label{tab:backbone_ablation}
\end{table*}

\subsection{Lightweight and Effective Prediction Head LFF}

Layer scale \cite{touvron2021going} is a technique used in Transformers, where multiple layers of self-attention and feed-forward networks are typically stacked, with each layer introducing a learnable scaling parameter $\gamma$. This scaling parameter can learn different values, enabling more effective information transfer throughout the network. Currently, feature fusion methods are usually implemented through simple operations like addition or concatenation \cite{lin2017feature}, which only provide fixed linear aggregation of feature maps without considering whether this combination is optimal for specific objects. For the model’s final prediction, our goal is to design a simple yet effective prediction head. Inspired by the Layer scale mechanism in Transformer architecture, we introduce a learnable parameter for each feature map to control the scaling ratio, allowing for more adaptive feature fusion.

\begin{figure}[t]
\centering
\includegraphics[width=0.98\columnwidth]{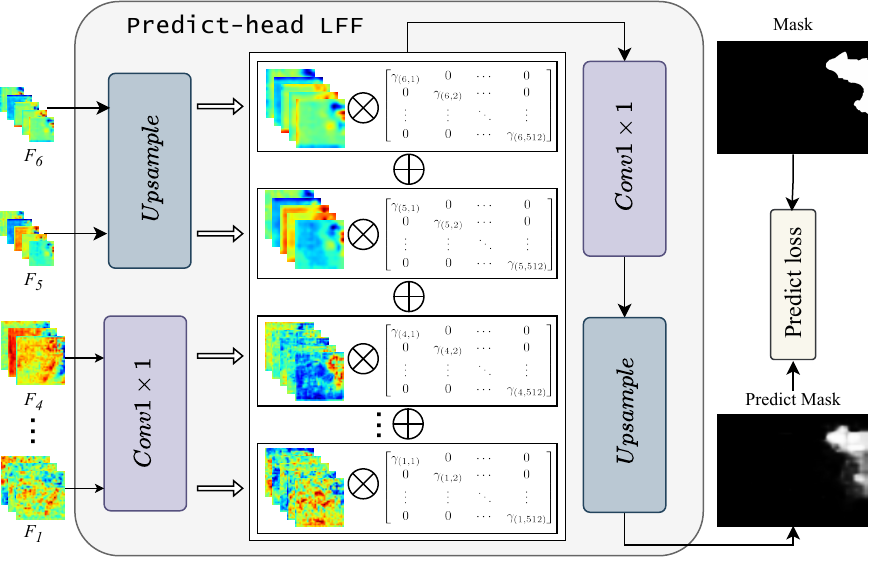} 
\caption{The Structure of LFF. By introducing learnable parameters $\gamma$, LFF dynamically adjusts the contribution of each feature map channel to the fusion result.}
\label{lff}
\end{figure}

The proposed LFF (Learnable Feature Fusion) prediction head is composed of five main parts, as shown in Figure \ref{lff}. First, the channels of feature maps $F_1$ to $F_4$ are unified to 512 dimensions using an LFF layer. Feature maps $F_5$ and $F_6$ are upsampled to one sixteenth of the original size. Then, each feature map is multiplied by its corresponding $\gamma$ scaling parameter, which is initialized to a small value like 1e-6. After that, all scaled feature maps are summed using another LFF layer, and the channel dimension of the summed result is reduced to 1. Finally, the result is upsampled, and the upsampled $H \times W \times 1$ mask is used as the final prediction result. The LFF process can be formalized as follows:
\begin{equation}
F_i = \text{Linear}(C_i, C)(F_i), \quad i = 1 \ldots 4
\end{equation}
\begin{equation}
F_i = \text{Upsample}\left(\frac{H}{16} \times \frac{W}{16}\right)(F_i), \quad i = 5, 6
\end{equation}
\begin{equation}
M_p = \text{Add}\left(F_i \times \gamma \right), \quad i = 1 \ldots 6
\end{equation}
\begin{equation}
M_p = \text{Linear}(C, 1)(M_p)
\end{equation}
\begin{equation}
M_p = \text{Upsample}\left(H \times W \right)(M_p)
\end{equation}

By setting the feature map weight parameters, the model can dynamically adjust each feature map's contribution to the fusion result, thereby enhancing the flexibility of feature fusion. Through this simple design, the model can better balance and integrate multi-scale features, highlighting important features while suppressing irrelevant or redundant ones.

\begin{table}[t]
\centering
\setlength{\tabcolsep}{2pt} 
\begin{tabular}{@{}ccccc@{}}
    \toprule
    \multirow{2}{*}{\textbf{Version}} & \multicolumn{4}{c}{\textbf{Pixel-level F1}}                                                         \\ \cmidrule(l){2-5} 
                                    & \textbf{COVERAGE} & \textbf{Columbia} & \textbf{CASIAv1} & \textbf{NIST16}  \\ \midrule
    Single Scale                      & 0.485                 & 0.936                 & 0.813                & 0.355                           \\
    MLP                               & 0.492                 & 0.955                 & 0.818                & 0.372                             \\
    LFF                               & \textbf{0.513}                 & \textbf{0.959}                 & \textbf{0.827}               & \textbf{0.384}                              \\ \bottomrule
    \end{tabular}
\caption{Comparison of Multi-Scale Feature Fusion. The best numbers in each column are highlighted in bold.}
\label{tab:muliscale_ablation}
\end{table}

\begin{table*}[t]
\centering
\setlength{\tabcolsep}{2pt}
\begin{tabular}{@{}ccccccccccc@{}}
\toprule
\multirow{2}{*}{\textbf{Method}}  & \multicolumn{5}{c}{\textbf{Pixel-level F1}}                              & \multicolumn{5}{c}{\textbf{Pixel-level AUC}}                                              \\ \cmidrule(r){2-6} \cmidrule(l){7-11}  
      & \textbf{COVERAGE} & \textbf{Columbia} & \textbf{CASIAv1} & \textbf{NIST16} & \textbf{AVG} & \textbf{COVERAGE} & \textbf{Columbia} & \textbf{CASIAv1} & \textbf{NIST16} & \textbf{AVG} \\ \midrule
ManTraNet                               & 0.196             & 0.462             & 0.327            & 0.193           & 0.295            & 0.566                 & 0.724                 & 0.643                & 0.709               & 0.661            \\
PSCC-Net                            & 0.379             & 0.864             & 0.592            & \underline{0.369}           & 0.551            & 0.884             & 0.946             & 0.893            & 0.828           & 0.888            \\
MVSS                          & 0.439             & 0.740              & 0.583            & 0.348           & 0.528            & 0.845             & 0.934             & 0.915            & 0.792           & 0.872            \\
CAT-Net                                                    & 0.428             & \underline{0.915}             & 0.808            & 0.252           & 0.601            & \underline{0.921}             & 0.946             & \underline{0.978}            & 0.824           & \underline{0.917}            \\
TruFor                       & \underline{0.457}             & 0.885             & \underline{0.818}            & 0.348           & \underline{0.627}            & 0.846             & \textbf{0.992}             & 0.897            & \underline{0.845}           & 0.895            \\
Ours (SparseViT)                            & \textbf{0.513}             & \textbf{0.959}             & \textbf{0.827}            & \textbf{0.384}            & \textbf{0.671}            & \textbf{0.935}             & \underline{0.970}              & \textbf{0.982}            & \textbf{0.861}           & \textbf{0.937}            \\ \bottomrule
\end{tabular}
\caption{Pixel-level Performance. The results show the F1 scores calculated under a fixed threshold of 0.5 and the AUC values obtained using the optimal F1-weighted configuration. The top-ranked results are highlighted in bold, and the second-ranked results are underlined.}
\label{tab:pixel_f1_and_auc}
\end{table*}

\begin{figure}[t]
\centering
\includegraphics[width=0.98\columnwidth]{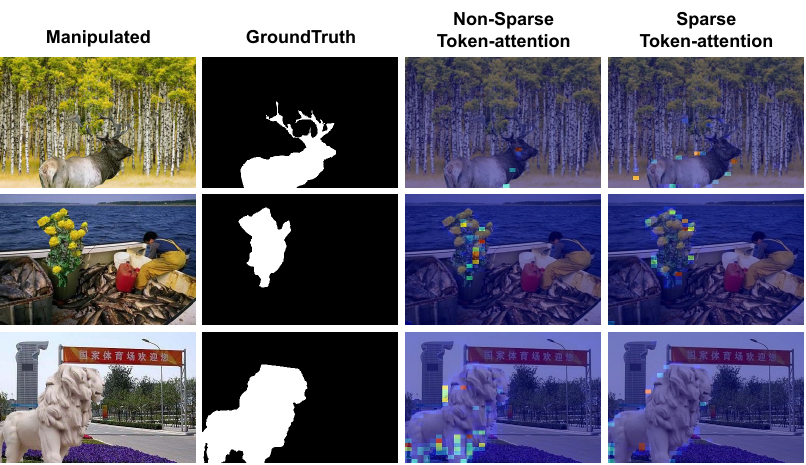} 
\caption{
We select an anchor point in the manipulation region and observe how other labels contribute to its attention. After sparsification, the anchor point’s attention focuses more on the manipulation-related edge regions containing non-semantic information, rather than on the surrounding semantic regions.}
\label{token_attention}
\end{figure}

\section{Results}
\subsection{Experimental Setup}
To ensure a fair comparison with existing state-of-the-art image manipulation localization methods, we trained our model on the dataset introduced by CAT-Net \cite{kwon2021cat} and then tested it on CASIAv1 \cite{CASIA_2013}, NIST16 \cite{NIST16_2019}, COVERAGE \cite{Coverage_2016}, Columbia \cite{Columbia_2006}, and DEF-12k \cite{defacto_2019} datasets. Similar to most previous works \cite{wei2023secondary} \cite{ma2024imdl}, we used pixel-level F1 scores and AUC (Area Under the Curve) to measure the model's performance. Unless otherwise specified, we reported results using a default threshold of 0.5. For detailed information on the experimental setup and the DEF-12k dataset, refer to Appendix A.

\subsection{Ablation Studies}

To better assess the performance impact of each component, we adopt an incremental approach by gradually adding components and comparing them with the full model that includes all components. This method allows us to thoroughly measure and optimize the architecture of our proposed model. We examine the effects of using sparse attention versus global attention on model parameters and floating-point operations (FLOPs). Additionally, we compare the capability of manually designed feature extractors and sparse attention mechanisms in extracting non-semantic features.
To explore the impact of the LFF prediction head, we compared its performance with the MLP prediction head from SegFormer \cite{xie2021segformer} under the introduction of sparse attention. This comparison not only helped us assess the effectiveness of the prediction head design but also revealed the specific impact of different heads on overall model performance. Additionally, we compared the traditional single-scale supervision with our proposed multi-scale supervision method to investigate the advantages of multi-scale supervision and its contribution to model performance.
The results for all these evaluations are reported based on training conducted on the dataset proposed by CAT-Net and tested on CASIAv1, NIST16, COVERAGE, Columbia, and DEF-12k. The experimental results are shown in Table \ref{tab:backbone_ablation} and Table \ref{tab:muliscale_ablation}.

\begin{figure*}[t]
\centering
\includegraphics[width=\textwidth]{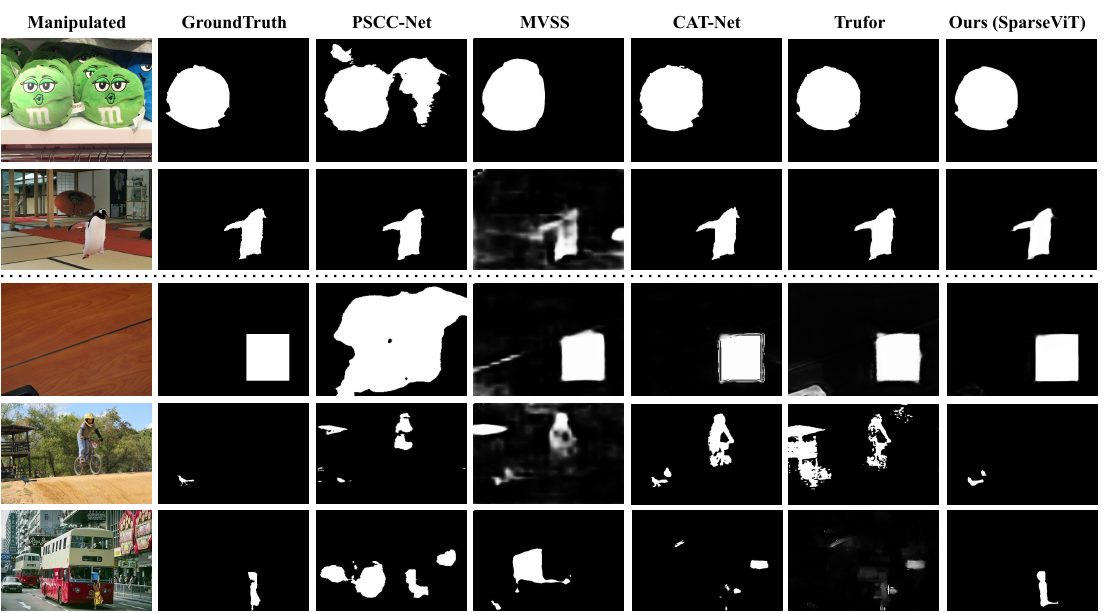}
\caption{IML by the SoTA. Existing models exhibit noticeable semantic-related false positives in the last three rows. Our model, SparseViT, effectively ignores semantic-related distractions through its unique sparse self-attention mechanism, focusing on capturing features that are unrelated to the semantic content but crucial to the integrity of the image.}
\label{model_compare}
\end{figure*}

\textbf{Sparse attention is effective in capturing non-semantic information.} In Table \ref{tab:backbone_ablation}, we compared the performance of sparse attention and global attention across five datasets. Additionally, we reported the performance of manually extracted non-semantic features and sparse attention on these five datasets. The results consistently confirmed the significant advantage of the sparse attention mechanism in extracting non-semantic features from manipulated images. We observed that certain handcrafted feature extraction methods did not significantly enhance model performance on the datasets, and in some cases, even led to performance degradation. This raises questions about the effectiveness of manual non-semantic feature extraction, warranting further investigation. However, it is evident that the sparse attention mechanism significantly improves model performance across all datasets, achieving comprehensive enhancements on five different datasets. 

Moreover, the design of sparse attention also shows its advantage in reducing computational burden. Compared to global attention, sparse attention reduces the model's floating-point operations by approximately 15\%, which is especially valuable in large-scale image processing tasks. In summary,  sparse attention enhances the model's sensitivity to subtle artifacts by precisely extracting non-semantic information in manipulated images, thereby significantly improving the model's generalization ability. 

As shown in Figure \ref{token_attention}, we qualitatively demonstrate that after sparsification, the model successfully suppresses semantic features that require dense encoding and long-range context dependencies, while being able to extract non-semantic features that do not require dense encoding. In the Appendix C, we conduct a qualitative analysis of sparse attention and handcrafted feature extractors.

\textbf{Influence of LFF.} In Table \ref{tab:muliscale_ablation}, we report the performance of single-scale features, LFF, and MLP \cite{xie2021segformer} prediction heads on the dataset. The experimental results show that regardless of using single-scale or multi-scale features, or adopting different feature fusion strategies, the F1 score on the CASIAv1 dataset exhibits high consistency. We attribute this phenomenon to the fact that CASIAv1 and CASIAv2 are sourced from the same dataset, thus the performance on the CASIAv1 dataset is not sufficient to reflect the model's generalization ability \cite{ma2023iml}. Further analysis reveals that both the LFF prediction head and the MLP prediction head achieve significant improvements in average F1 scores across the five datasets compared to using only single-scale features. This indicates that effective feature fusion strategies can significantly enhance the model's performance in detecting image manipulation. Specifically, the LFF also achieves an improvement in mean F1 compared to the MLP prediction head, validating that learnable feature fusion outperforms simple feature addition in terms of performance.

\begin{figure}[t]
\centering
\includegraphics[width=\columnwidth]{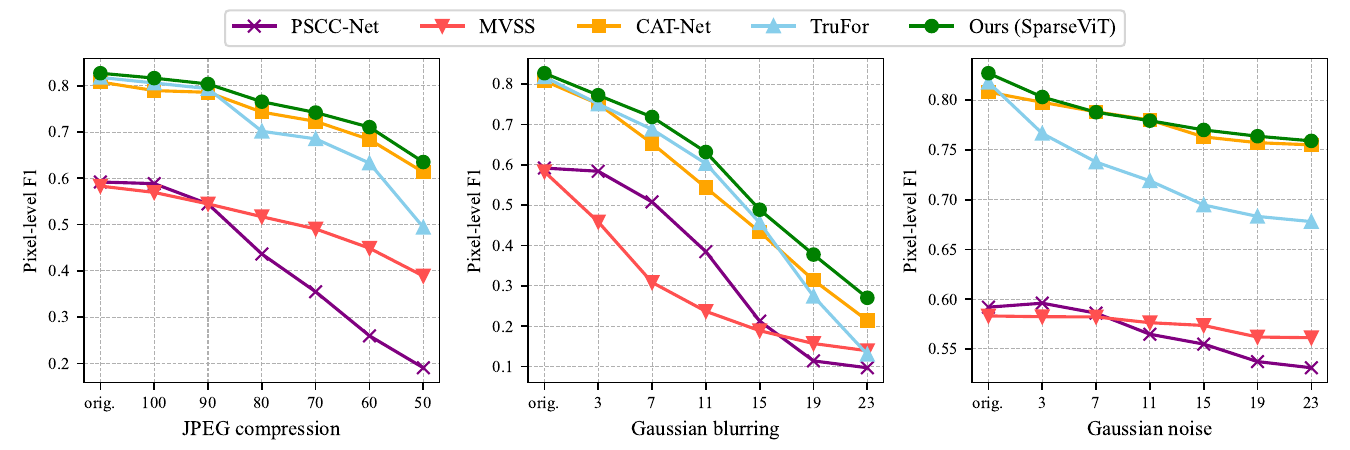} 
\caption{Robustness Analysis on CASIAv1. The results are presented in terms of pixel-level F1 scores.}
\label{robust}
\end{figure}
The advantage of LFF lies in its ability to adaptively learn the optimal fusion weights between different feature maps, rather than just adding them. This learning mechanism allows LFF to more precisely handle multi-scale features, thereby better capturing manipulation traces in images. Additionally, the use of multi-scale features has proven beneficial, as it provides different levels of semantic and non-semantic information, aiding the model in making more accurate predictions under various operational conditions.

\subsection{State-of-the-Art Comparison}
To ensure fairness in the evaluation, we only considered models whose code is publicly available online. We followed the same protocol as CAT-Net, retrained these models, and tested them on public datasets. In this study, we considered a variety of methods and ultimately included four approaches that rely on handcrafted extraction of non-semantic features for manipulated images: ManTraNet, MVSS, CAT-Net v2, and TruFor. Additionally, we included one method that does not use handcrafted feature extraction: PSCC-Net \cite{liu2022pscc}. A brief summary of these methods is provided in Table \ref{tab:model_information} for reference. Our goal is to provide a comprehensive and fair comparison to gain deeper insights into the performance and potential of different approaches in image manipulation localization. 

\begin{table}[t]
\centering
\begin{tabular}{@{}c|c|c|c@{}}
\toprule
\textbf{Method} & \textbf{Size} & \textbf{\begin{tabular}[c]{@{}c@{}}Parameter\end{tabular}} & \textbf{\begin{tabular}[c]{@{}c@{}}FLOPs\end{tabular}} \\ \midrule
ManTraNet       & 256$\times$256             & 3.9M                                                                & 274.0G                                                            \\
PSCC-Net        & 256$\times$256             & 3.7M                                                                & 45.7G                                                            \\
MVSS            & 512$\times$512             & 147.0M                                                                & 167.0G                                                            \\
CAT-Net         & 512$\times$512             & 114.0M                                                                & 134.0G                                                            \\
TruFor          &512$\times$512             & 68.7M                                                                & 236.5G                                                            \\
Ours (SparseViT)            & 512$\times$512             & 50.3M                                                                & 46.2G                                                            \\ \bottomrule
\end{tabular}
\caption{Comparison with the State-of-the-Art on Parameter and FLOPs.}
\label{tab:best_param}
\end{table}

\textbf{Localization results.} In Table \ref{tab:pixel_f1_and_auc}, we present the performance of various methods in pixel-level localization. Our method stands out with its superior average F1 scores, ranking best across all datasets. A detailed analysis of these results reveals that our model outperforms both traditional methods based on handcrafted non-semantic feature extraction and models that do not rely on handcrafted features. The reason our model excels among many others lies in its innovations in feature learning and representation. By deeply exploring the intrinsic structures of manipulated images, our model can accurately capture the subtle traces left by manipulation. Even when facing complex and varied manipulation techniques, it maintains high accuracy in detection.

\textbf{Detection results.} 
We selected the weight parameters that performed best in terms of the Pixel-F1 metric to evaluate the model's AUC performance. By analyzing the data in Table \ref{tab:pixel_f1_and_auc}, we observe that our SparseViT model achieved the best performance across nearly all tested datasets and exhibited the highest average AUC value. This result indicates that the SparseViT model outperforms existing baselines across a broad range of performance evaluation points.

\textbf{Comparison of model size.} Compared to the current top-performing Trufor, SparseViT not only achieves superior F1 and AUC performance with the same training data size (512$\times$512 pixels) but also reduces the model size by over 80\%. Additionally, even when compared to the ManTraNet, which uses smaller training data (256$\times$256 pixels), SparseViT shows a significant advantage in reducing computational load. The specific data is shown in Table \ref{tab:best_param}.

\textbf{Robustness analysis.} Following the guidelines of references \cite{wu2019mantra} and \cite{hu2020span}, we evaluated the robustness of the model against three common attack methods in image manipulation localization on the CASIAv1 dataset, namely JPEG compression, Gaussian blur, and Gaussian noise. The results are shown in Figure \ref{robust}. Observations indicate that SparseViT outperforms existing state-of-the-art models in resisting these disturbances, demonstrating superior robustness.

Overall, compared to existing models tested under a fair cross-dataset evaluation protocol, our model achieves state-of-the-art performance. Figure \ref{model_compare} qualitatively illustrates a key advantage of our model: regardless of whether object-level manipulation is involved, our model effectively utilizes non-semantic features that are independent of the image's semantic content to accurately identify manipulated regions, thereby avoiding semantic-related false positives.

\section{Conclusions}
Relying on handcrafted methods to enhance a model's ability to extract non-semantic features often limits its generalization potential in unfamiliar scenarios. To move beyond manual approaches, we propose using a sparse self-attention mechanism to learn non-semantic features. Sparse self-attention directs the model to focus more on manipulation-sensitive non-semantic features while suppressing the expression of semantic information. Our adaptive method is not only parameter-efficient but also more effective than previous handcrafted approaches, with extensive experiments demonstrating that SparseViT achieves SoTA performance and generalization ability.

\appendix
\section{Acknowledgments}
This work was jointly supported by the Sichuan Natural Science Foundation under grant 2024YFHZ0355, Sichuan Major Projects under grant 2024ZDZX0001, and the Science and Technology Development Fund, Macau SAR, under grants 0141/2023/RIA2 and 0193/2023/RIA3.  Numerical computations were jointly supported by Hefei Advanced Computing Center and Chengdu Haiguang Integrated Circuit Design Co., ltd. with HYGON K100AI DCU units.

\bibliography{aaai25}
\appendix
\newpage
\section{Appendix}
\subsection{Appendix A. Details of the Experimental Setup}
\textbf{Datasets.} To ensure a fair comparison with current state-of-the-art Image Manipulation Localization (IML) methods, our model was trained on a dataset provided by CAT-Net \cite{kwon2021cat}. Subsequently, we tested the trained model on widely recognized public datasets in the image manipulation localization field. These datasets include CASIAv1 \cite{CASIA_2013}, NIST16 \cite{NIST16_2019}, COVERAGE\cite{Coverage_2016}, Columbia\cite{Columbia_2006}, and DEFACTO \cite{defacto_2019}. Specifically, given that the DEFACTO dataset lacks real images as negative samples, we employed the approach proposed by MVSS \cite{dong2022mvss} to address this issue. We randomly selected 6,000 images from the DEFACTO dataset as positive samples and similarly extracted 6,000 images from the MS-COCO dataset as negative samples. These 12,000 images collectively form our DEF-12k dataset for testing. This approach ensures that during evaluation, the model not only demonstrates its performance on diverse datasets but also undergoes effective testing even in the absence of standard negative samples.

\textbf{Evaluation Criteria.} In our evaluation process, as with most previous studies, we used pixel-level F1 score and AUC (Area Under the Curve) as key metrics to measure model performance. We acknowledge that using the optimal threshold for evaluation may lead to overly optimistic performance estimates, as the ideal threshold is often unknown in real-world applications and may vary across different scenarios. To avoid this and provide a more practical and comparable performance assessment, we employed a fixed threshold in the evaluation report unless otherwise specified. Specifically, we chose 0.5 as the default threshold for reporting the model's performance metrics.

\textbf{Implementation.} Our SparseViT model was carefully implemented in the PyTorch framework and efficiently trained on an NVIDIA RTX 3090 GPU. During training, we selected a batch size of 16 and set 200 training epochs to ensure that the model could fully learn and converge. For optimization, we used the Adam optimizer with an initial learning rate of $1 \times 10^{-4}$ , which was then periodically decayed to $1 \times 10^{-7}$ using a cosine annealing strategy. This approach helps the model to finely approach the optimal solution during training. Similar to MVSS-Net, we performed data augmentation before training to enhance the model's generalization capability. The data augmentation techniques used include image flipping, blurring, compression, and simple manipulation operations, which help to simulate various transformations and manipulations that images may undergo in the real world. Additionally, to further improve the model's performance, we employed a pre-training strategy. Specifically, we initialized our SparseViT model using the Uniformer \cite{li2023uniformer} weights pre-trained on the ImageNet-1k dataset.

\subsection{Appendix B. The Combination of Sparsity rates}
\begin{table}[t]
\centering
\setlength{\tabcolsep}{2pt} 
\begin{tabular}{@{}ccccc@{}}
\toprule
\multirow{2}{*}{Sparsity rate} & \multicolumn{4}{c}{Pixel-level F1}     \\ \cmidrule(l){2-5} 
                               & COVERAGE & Columbia & CASIAv1 & NIST16 \\ \midrule
2                              & 0.461    & \textbf{0.967}    & \underline{0.819}   & 0.348  \\
4                              & 0.479    & 0.945    & 0.813   & 0.357  \\
8                              & \underline{0.484}    & 0.949    & 0.810   & \underline{0.368}  \\ \midrule
SparseViT                      & \textbf{0.513}    & \underline{0.959}    & \textbf{0.827}   & \textbf{0.384}  \\ \bottomrule
\end{tabular}
\caption{ Pixel-level F1 Performance with Different Sparsity rates. All single sparsity rates utilize the LFF prediction head.}
\label{tab:compare_sparse}
\end{table}


Although we have introduced the hyperparameter ``sparsity rate'' to achieve sparsity in global self-attention for extracting non-semantic features, different levels of sparsity in the attention mechanism can identify non-semantic features at varying degrees. Therefore, selecting the ``sparsity rate'' for our model becomes crucial for extracting non-semantic features. 

In our study, we conducted a series of experiments focusing on the combination of sparsity rates within the model. First, we explored the impact of a single sparsity rate on the extraction of non-semantic features. As shown in Table \ref{tab:compare_sparse}, we tested the model's pixel-level F1 scores under different sparsity rates (2, 4, 8) across four different datasets. The experimental results indicate that on the CASIAv1 and Columbia datasets, models with lower sparsity rates achieved similar or even higher F1 scores compared to those with higher sparsity rates, whereas their performance on the NIST16 and COVERAGE datasets was inferior to that of the high sparsity rate models. 

\begin{figure}[t]
\centering
\includegraphics[width=\columnwidth]{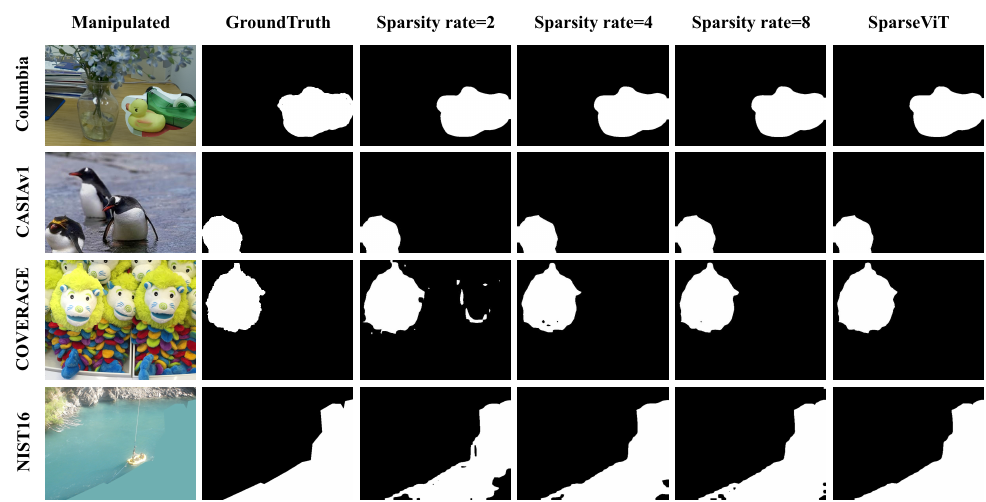}
\caption{Qualitative Analysis Under Different Sparsity rates. We randomly selected one image from each of the four datasets to demonstrate the localization capability of different sparsity rates on these datasets.}
\label{qutative_sparsity}
\end{figure}

\begin{figure*}[t]
\centering
\includegraphics[width=\textwidth]{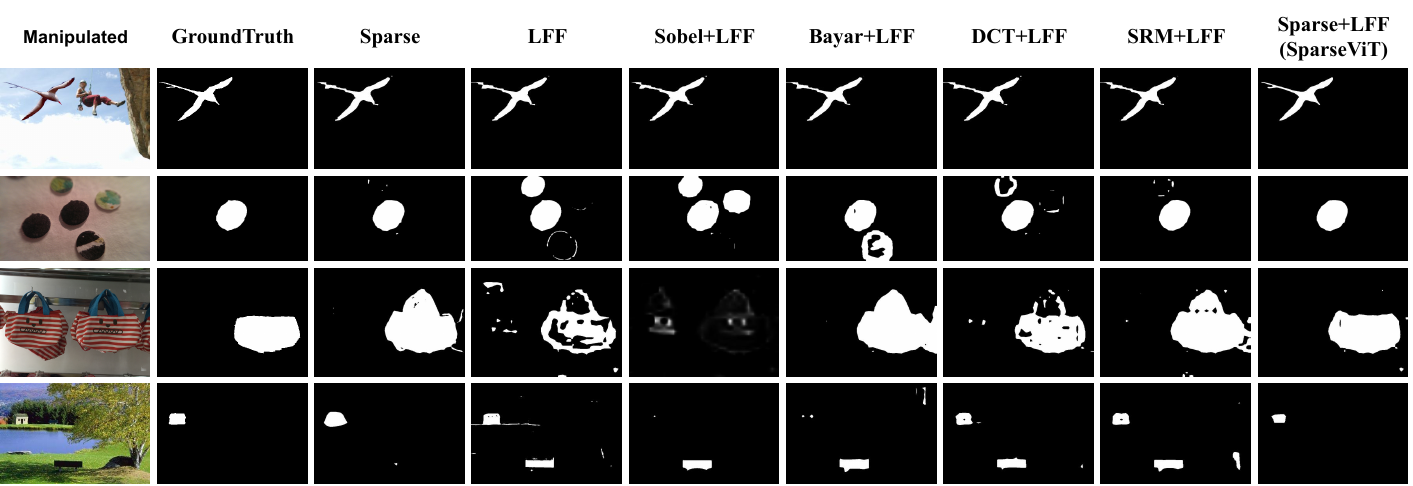}
\caption{Examples of the ability of sparse self-attention and handcrafted feature extractors to localize manipulated regions.}
\label{qutative_extractor}
\end{figure*}
Our analysis revealed that lower sparsity rates are less effective in suppressing semantic information compared to higher sparsity rates. This suggests that on datasets like CASIAv1 and Columbia, which contain more object-level manipulations, the model can still achieve good F1 scores even if it learns incorrect semantic associations. However, on meticulously designed datasets like NIST16 and COVERAGE, the model's generalization ability is limited due to insufficient learning of non-semantic features. In Figure \ref{qutative_sparsity}, we conducted a qualitative analysis of different sparsity levels. As revealed by the F1 scores, models with lower sparsity rates underperformed in resisting semantic associations due to insufficient learning of non-semantic features. This resulted in poorer localization performance on high-quality datasets like NIST16 compared to models with higher sparsity rates.

To overcome this limitation and enhance the model's learning of non-semantic features while improving its generalization ability, we propose a new strategy: applying sparsification to self-attention with exponentially decreasing sparsity rates across different layers of the model. This approach aims to balance the model's learning of both non-semantic and semantic features, enabling the model to remain sensitive to non-semantic features while also capturing some semantic information, thereby achieving more balanced and robust performance across various datasets.

\subsection{Appendix C. Qualitative Comparison Results}

In Figure \ref{qutative_extractor}, we compare the ability of handcrafted feature extractors and the sparse self-attention method to localize manipulated regions in images. The results show that the DCT and SRM handcrafted feature extractors achieve some improvement in identifying manipulated areas. However, the Sobel and Bayar feature extractors, when combined with the LFF prediction head, do not surpass the localization performance of using the LFF prediction head alone. This raises the question of whether all handcrafted feature extractors can effectively extract non-semantic features from images. It is evident that the sparse self-attention mechanism, even without relying on the LFF prediction head, demonstrates superior localization ability compared to DCT and other handcrafted feature extractors. This finding confirms the capability of sparse self-attention in adaptively extracting non-semantic features from manipulated images, suggesting that, compared to traditional handcrafted methods, the sparse self-attention mechanism might be more effective in capturing non-semantic information within images.

\begin{table}[t]
\centering
\begin{tabular}{@{}cccc@{}}
\toprule
\textbf{Method} & \textbf{AVG F1} & \textbf{Parameter} & \textbf{FLOPs} \\ \midrule
ASPP            & 0.647             & 18.35M             & 5.46G          \\
AFF             & 0.669             & 3.67M              & 2.06G          \\
LFF             & 0.671             & 0.66M              & 0.68G          \\ \bottomrule
\end{tabular}
\caption{The parameter efficiency and performance of different fusion techniques.}
\label{tab:compare_ff}
\end{table}

\begin{table*}[t]
\centering
\begin{tabular}{@{}cccccc@{}}
\toprule
\multirow{2}{*}{\textbf{Method}} & \multicolumn{5}{c}{\textbf{Pixel-level IoU}}                                              \\ \cmidrule(l){2-6} 
                                 & \textbf{COVERAGE} & \textbf{Columbia} & \textbf{CASIAv1} & \textbf{NIST16} & \textbf{AVG} \\ \midrule
PSCC-Net                         & 0.301             & 0.814             & 0.459            & 0.294           & 0.467            \\
MVSS                             & 0.371             & 0.658             & 0.505            & 0.269           & 0.451            \\
CAT-Net                          & 0.388             & \underline{0.895}             & 0.754            & 0.213           & 0.563            \\
TruFor                           & \underline{0.415}             & 0.859             & \underline{0.764}            & \underline{0.301}           & \underline{0.585}            \\
Ours (SparseViT)                 & \textbf{0.472}             & \textbf{0.938}             & \textbf{0.775}            & \textbf{0.331}           & \textbf{0.629}            \\ \bottomrule
\end{tabular}
\caption{ Pixel-level IoU. Consistent with the Pixel-level F1 results, SparseViT achieves the best performance in Pixel-level IoU compared to current state-of-the-art models.}
\label{tab:pixel-level IoU}
\end{table*}

\subsection{Appendix D. IoU Results Report}
We report the Pixel-level IoU scores of the most advanced IML models, as shown in Table \ref{tab:pixel-level IoU}. SparseViT achieves the best results across all four datasets. Not only does SparseViT excel in pixel-level F1, but it also demonstrates high precision and robustness in overall image segmentation and recognition tasks. This is attributed to SparseViT’s unique sparse structure design, which significantly enhances the model’s ability to capture non-semantic features while maintaining parameter efficiency. 

\subsection{Appendix E. Implement sparse coding on other ViTs}

We chose Uniformer because models like PVT \cite{wang2021pyramid} and Segformer \cite{xie2021segformer} use overlapping patch partitioning, which may make sparse interactions between patches less controllable and lead to overfitting semantics. Additionally, Uniformer uses CNNs in the shallow layers to extract features, and we believe that CNN's ability to capture basic features, such as edges, is beneficial for IML. Our approach is also compatible with vanilla ViT \cite{dosovitskiy2020image16}, as shown in the Table \ref{tab:other vit}. We implemented sparse attention on vanilla ViT and VOLO \cite{yuan2022volo} (without LFF), and the results demonstrate that our method is equally effective for vanilla ViT.

\begin{table*}[t]
\centering
\begin{tabular}{@{}ccccccc@{}}
\toprule
\multicolumn{2}{c}{\multirow{2}{*}{\textbf{Method}}} & \multicolumn{5}{c}{\textbf{Pixel-level F1}}                                               \\ \cmidrule(l){3-7} 
\multicolumn{2}{c}{}                                 & \textbf{COVERAGE} & \textbf{Columbia} & \textbf{CASIAv1} & \textbf{NIST16} & \textbf{AVG} \\ \midrule
\multirow{2}{*}{ViT}           & non-sparse          & 0.417             & 0.858             & 0.693            & 0.345           & 0.578        \\
                               & sparse              & 0.441             & 0.872             & 0.708            & 0.372           & 0.598        \\ \midrule
\multirow{2}{*}{VOLO}          & non-sparse          & 0.382             & 0.776             & 0.646            & 0.235           & 0.510        \\
                               & sparse              & 0.404             & 0.779             & 0.645            & 0.259           & 0.522        \\ \bottomrule
\end{tabular}
\caption{ Implement sparse coding on vanilla ViT and VOLO to validate the effectiveness of the proposed sparsification method.}
\label{tab:other vit}
\end{table*}

\subsection{Appendix F. The role of LFF in improving performance}
One of the goals of designing LFF is to achieve both lightweight and efficient performance. Therefore, in the ``Influence of LFF'' section, we focus on comparing it with MLP, which is designed for lightweight purposes. To further highlight the advantages of LFF in terms of both lightweight design and efficiency, we provide additional comparisons with AFF \cite{dai2021attentional} and ASPP \cite{chen2017deeplab} in Table \ref{tab:compare_ff}. The results demonstrate that SparseViT outperforms these methods in terms of average F1 score and parameter efficiency, proving that LFF can significantly reduce model complexity and computational cost while maintaining performance.

\end{document}